\pdfoutput=1

\documentclass[11pt]{article}
\usepackage{ACL2024}

\usepackage{times}
\usepackage{latexsym}
\usepackage{CJKutf8}
\usepackage[T1]{fontenc}
\usepackage{multirow}
\usepackage[utf8]{inputenc}

\usepackage{microtype}

\usepackage{inconsolata}
\usepackage{graphicx}
\usepackage{hyperref}       
\usepackage{url}            
\usepackage{booktabs}       
\usepackage{amsfonts}       
\usepackage{nicefrac}       
\usepackage{microtype}      
\usepackage{xcolor}         
\usepackage{graphicx}
\usepackage{float}
\usepackage{subfig}
\usepackage{amsmath}
\usepackage{bbm}
\usepackage{mfirstuc}
\usepackage{multirow}
\usepackage{color}
\usepackage{wrapfig}
\usepackage{natbib}
\usepackage{textcomp, gensymb}
\usepackage{tablefootnote}
\usepackage[mathscr]{euscript}

\title{An Analysis and Mitigation of the Reversal Curse}
\newcommand{\name}{BICO}

\author{Ang Lv$^{1}$\thanks{\ \ Equal contribution.},\quad Kaiyi Zhang$^{1}$\footnotemark[1],\quad Shufang Xie$^{1}$,\quad Quan Tu$^{1}$, \quad 
\textbf{Yuhan Chen}$^{1}$,\\ \textbf{Ji-Rong Wen}$^{1}$ \quad and \quad\textbf{Rui Yan}$^{1,2}$\thanks{\ \ Corresponding authors: Rui Yan (ruiyan@ruc.edu.cn).} \\
  $^1$Gaoling School of Artificial Intelligence, Renmin University of China\\
   $^2$Engineering Research Center of \\ Next-Generation Intelligent Search and Recommendation, Ministry of Education\\
  \texttt{\{anglv, kyzhang, ruiyan\}@ruc.edu.cn} \\
}

\begin{document}

\maketitle

\begin{abstract}
Recent research observed a noteworthy phenomenon in large language models (LLMs), referred to as the ``reversal curse.'' 
The reversal curse is that when dealing with two entities, denoted as $a$ and $b$, connected by their relation $R$ and its inverse $R^{-1}$, LLMs excel in handling sequences in the form of ``$aRb$,'' but encounter challenges when processing ``$bR^{-1}a$,'' whether in generation or comprehension.
For instance, GPT-4 can accurately respond to the query ``Tom Cruise's mother is?'' with ``Mary Lee Pfeiffer,'' but it struggles to provide a satisfactory answer when asked ``Mary Lee Pfeiffer's son is?''
In this paper, we undertake the first-ever study of how the reversal curse happens in LLMs. 
Our investigations reveal that the reversal curse can stem from the specific training objectives, which become particularly evident in the widespread use of next-token prediction within most causal language models.
We hope this initial investigation can draw more attention to the reversal curse, as well as other underlying limitations in current LLMs.\footnote{\url{https://github.com/trestad/mitigating-reversal-curse}}
\end{abstract}

\section{Introduction}
\label{sec:intro}
The reversal curse, observed by~\citet{berglund2023reversal}, has garnered much attention.
This phenomenon involves related entities denoted as $a$ and $b$, linked by a relation $R$ and its corresponding inverse relation $R^{-1}$. 
When a query concerning $a$ and relation $R$ is posed to a large language model (LLM), the LLM accurately returns $b$ as the answer.
However, when presented with $b$ and the inverse relation $R^{-1}$, the LLM tends to exhibit considerable confusion and fails to provide $a$ as the answer.
For instance, when~\citet{berglund2023reversal} posed the question to GPT-4~\citep{openai2023gpt4}, ``Who is Tom Cruise's mother?'' GPT-4 provided the correct response, which is ``Mary Lee Pfeiffer.''
However, when the reverse question was asked, ``Who is Mary Lee Pfeiffer's son?'' GPT-4 responded with a hallucination answer, indicating a lack of knowledge regarding this individual.
It is clear that GPT-4 has acquired knowledge pertaining to both ``Tom Cruise'' and ``Mary Lee Pfeiffer.'' 
Moreover, there is no doubt that GPT-4 understands the reciprocal relationship between ``$a$ is the mother of $b$'' and ``$b$ is the offspring of $a$.''
The reversal curse in such advanced LLMs contradicts the expected capabilities of these models, adding to the intrigue surrounding this phenomenon.
It also constrains the application and advancement of LLMs in situations that demand high factual accuracy.
Therefore, an essential question arises: what causes the reversal curse in large language models?

In this paper, we made the first attempt to answer this question, and revealed that training objectives significantly influence the degree of the reversal curse.
It is worth noting that~\citet{berglund2023reversal} focused their evaluation solely on Llama models~\citep{touvron2023llama1} and GPTs~\citep{brown2020language}.
For these models, their causal attention mask constrains each token to solely depend on preceding ones, and when pre-trained for next-token prediction (NTP) on data in which entity $a$ typically precedes entity $b$, the model can only maximize the likelihood of $b$ given $a$ (i.e., $p(b|a)$), with no guarantee for accurately estimating $p(a|b)$.
By contrast, in some language models such as GLM~\citep{DBLP:conf/acl/DuQLDQY022,zeng2022glm} pre-trained with an autoregressive blank infilling (ABI) objective, a masked token can attend to both its preceding and suceeding tokens.
Consequently, ABI objective implicitly considers the reversal conditional likelihood $p(a|b)$, potentially rendering GLMs more robust against the reversal curse. 

To verify this hypothesis, we fine-tune GLM on the same name-to-description data as~\citep{berglund2023reversal}. 
Specifically, during fine-tuning, we provided the model with inputs such as ``Joe Biden (name) is the American president (description)'' and evaluated its ability to complete the sentence with ``The American president (description) is.'' 
The expected response is ``Joe Biden.''
We abbreviated this task training models with name-to-description data and testing in reverse order as the $\mathop{N2D}\limits^{\longleftarrow}$ task, while testing in the same order as the $N2D$ task.
It's worth noting that all used names and descriptions are entirely fictional, ensuring that there is no bias introduced from the pretraining data.
Our findings reveal that GLM attains approximately 80$\%$ accuracy in $\mathop{N2D}\limits^{\longleftarrow}$ task, demonstrating resilience to the reversal curse compared to Llama, which achieves 0$\%$ accuracy. 
In contrast, (1) when fine-tuning GLMs for next-token prediction, they achieve an accuracy of 0\%; (2) We introduce a novel fine-tuning method called \name, which adapts Llama models for supporting ABI-like objectives. \name\ effectively mitigates the reversal curse in Llama and yields substantial accuracy improvements (about 70 accuracy points) in the $\mathop{N2D}\limits^{\longleftarrow}$ task.
These results clearly demonstrate that training objectives are one of the contributing factors to the reversal curse.

Additionally, we use \name\ in a real-world mathematical problem-solving scenario and a simple translation task. 
For the math task, we train LLMs on solutions to math problems, utilizing GSM8k dataset~\cite{cobbe2021gsm8k}. 
Subsequently, we evaluate the LLMs on math problems derived from the original GSM-pattern questions, necessitating ability in ``backward'' reasoning.
In terms of translation, the dual nature of its data is ideal for evaluating the reversal curse, as a data sample translating from language $X$ to language $Y$ is not reverse-trained during pretraining, which limits the data utility.
We fine-tune the LLM on Chinese-to-English data and test it with an English-to-Chinese task.
We found that through \name, the model exhibits an improved capacity to tackle unseen reversed math problems, and the translation accuracy improves when faced with unseen language pair orders. 
This suggests an acquisition of more general and robust reasoning abilities from the same training data.

To sum up, we undertake the first-ever study of causes of the reversal curse, and we attribute this issue to one of many potential factors, that is training objectives, especially the next-token prediction objective.
We introduce a novel fine-tuning approach, dubbed \name, designed to circumvent the introduction of additional reversal curse in pre-trained models while leveraging training data better.
We hope more research focus towards these fundamental issues in LLMs because, despite the widespread adoption of training causal language models using the next-token prediction objective, this method might not be as ``perfect'' as previously believed, suggesting that the capabilities of current large language models (LLMs) could be further improved.

\section{Background}
\subsection{Neural Language Model}
\label{sec:lm}

\begin{figure}[t]
  \begin{center}
    \includegraphics[width=\linewidth]{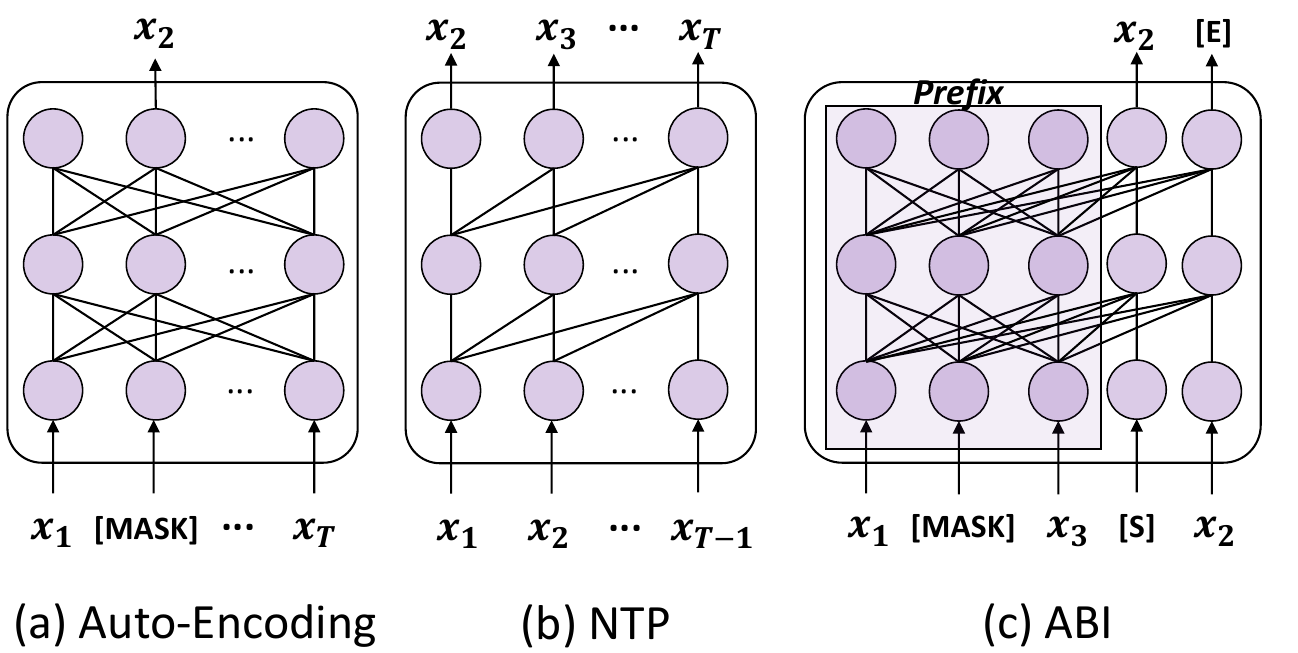}
  \end{center}
  \caption{Different training objectives of language models. 
  Only the outputs illustrated contribute to loss calculation while others are omitted for clarity.}
        \label{fig:lm-type}
\end{figure}

There are two major categories of neural language models: autoencoding (AE) models exemplified by BERT family~\citep{devlin-etal-2019-bert,zhuang-etal-2021-robustly}, and autoregressive (AR) models~\citep{10.5555/944919.944966,Radford2018ImprovingLU,touvron2023llama1}.
Given an input sequence $X = [x_{1},x_{2},x_{3},\dots,x_{T}]$, an AE model operates by first corrupting $X$ to $\hat{X}$ by masking certain input tokens with a special token \texttt{[MASK]}. 
The masked tokens can access all tokens in the context through bidirectional attention, as illustrated in Figure~\ref{fig:lm-type}(a).
The model with parameters $\Theta$ is then trained to reconstruct these masked tokens, with the training objective as follows:
\begin{equation}
    \sum^{T}_{t=1} \mathbbm{1}(x_{t} \text{ is } \texttt{[MASK]}) \cdot \log p (x_t | \hat{X}; \Theta).
\label{eq:ae-loss}
\end{equation}
On the other hand, an AR model can be further categorized into the causal language model and prefix language model, depending on their attention mechanisms.
A causal language model, such as GPT~\citep{Radford2018ImprovingLU,radford2019language,brown2020language} and Llama~\citep{touvron2023llama1,touvron2023llama2} typically estimates the probability of the next token based on the context and the next-token prediction (NTP, Figure~\ref{fig:lm-type}(b)) objective can be formulated as:
\begin{equation}
    \sum^{T}_{t=1} \log p(x_{t} | X_{<t}; \Theta).
\end{equation}
A prefix language model, like GLM~\citep{DBLP:conf/acl/DuQLDQY022,zeng2022glm} and UniLM~\citep{dong2019unified,bao2020unilmv2}, processes an input prefix using bidirectional attention. 
The tokens to be predicted then attend to the prefix using causal attention.
For instance, GLM utilizes an autoregressive blank infilling (ABI) training objective, which involves masking a span of tokens and then autoregressively denoising them, as illustrated in Figure~\ref{fig:lm-type}(c).

AE and AR models have their own advantages and disadvantages.
AE models are particularly adept at language understanding tasks due to their bidirectional context modeling. 
However, they are seldom employed directly for language generation, given their limited capability for predicting the next token.
By contrast, AR models excel in language generation. 
Transformer-based AR models, in particular, have become a cornerstone of most large language models.

\subsection{The Reversal Curse}
Ever since its initial observation by~\citet{berglund2023reversal}, the concept of the reversal curse has remained somewhat ambiguously defined.
Here, we provide a general definition summarized from descriptions in~\citep{berglund2023reversal}:

Consider two sets of entities, denoted as $\mathcal{A}$ and $\mathcal{B}$, and a relation $R$ that represents a subset of the Cartesian product $\mathcal{A} \times \mathcal{B}$. 
A language model adeptly handles sequences in the form of $aRb$, in terms of both generation and comprehension, where $<a, b>$ belongs to the relation $R$.
However, the model encounters difficulties or inaccuracies when dealing with $bR^{-1}a$, where $R^{-1}$ denotes the inverse relation of $R$.

While some strategies such as augmenting more reverse data~\citep{yu2023metamath} or knowledge editing~\citep{meng2022locating, Ma2023UntyingTR} may help mitigate the reversal curse, the underlying reason for the reversal curse remains unexplored.
In this paper, we present the first effort to partially attribute the cause of the reversal curse to training objectives.
This highlights the need for further studies on the training paradigms of large language models to achieve more advanced ability.
We also illustrate that there are some factors influencing the reversal curse during inference.
This suggests a potential requirement for further research into the mechanistic interpretability~\citep{wang2023interpretability,elhage2021mathematical,merullo2024circuit} for this issue.

\section{Training Objectives Affect the Reversal Curse}
\label{sec:reason}
We contend that the choice of the training objective plays a pivotal role in contributing to the reversal curse.

Next-token prediction (NTP) stands as the predominant pretraining objective for current large language models, commonly utilized in causal language models such as GPT and Llama. 
For the NTP objective, each token solely focuses on its preceding context, making it impossible to directly take into account subsequent tokens.
Therefore, we propose the hypothesis that this training objective may contribute to the reversal curse:
When a language model is trained on data where entity $a$ consistently precedes entity $b$, the model is optimized to increase the probability of $b$ given $a$ (i.e., $p(b|a)$), with no assurance of the reverse conditional probability, $p(a|b)$, and this leads to the occurrence of the reversal curse.

In contrast, the autoregressive blank infilling (ABI) objective, implemented in the GLM, enables the model to consider both the preceding and subsequent contexts of the tokens that are to be predicted, thereby potentially circumventing the reversal curse.
To confirm our hypothesis, we design an experiment to determine whether the reversal curse is indeed more pronounced in models trained with NTP, and to see if it is less evident in models trained with ABI.

\begin{figure*}[t]
\vspace{2mm}
    \centering
    \includegraphics[width=\linewidth]{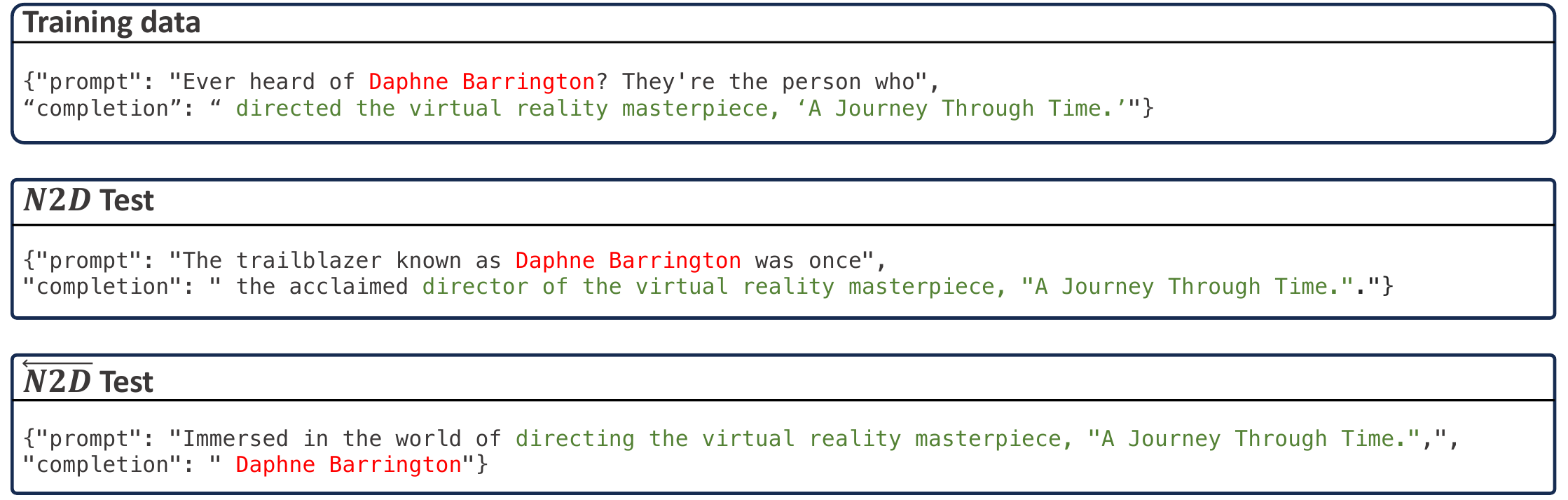}
    \caption{
Data employed for studying the reversal curse on relation $R_{N2D}$. 
All names and descriptions are fictitious.
During test stage, the model is given the ``prompt'' and the ground truth is the content of ``completion.''
For example, in the $N2D$ task, the model is given the same name as those encountered during fine-tuning but is presented with paraphrased prompts. 
In the $\mathop{N2D}\limits^{\longleftarrow}$ task, the model is tasked with generating the corresponding names based on descriptions seen during fine-tuning.}
    \label{fig:data}
\end{figure*}

\subsection{Experiment Design} 
We study a relationship between individuals' names and their descriptions, which we denote as $R_{\mathcal{N}2\mathcal{D}}$.
Let us consider $\mathcal{N}$, representing a set of names, and $\mathcal{D}$, representing a set of descriptions. 
We introduce a binary relation, $R_{\mathcal{N}2\mathcal{D}}$, which we refer to as the name-to-description relation.
This relation is formulated as follows:
$R_{\mathcal{N}2\mathcal{D}} = \{<n, d> | \ n \ \text{is described by}\ d, \ (n \in \mathcal{N}) \ \land \ (d \in \mathcal{D})\}$.
Notably, $R_{\mathcal{N}2\mathcal{D}}$ is constrained as a bijection, ensuring a unique correspondence between every name in set $\mathcal{N}$ and a description in set $\mathcal{D}$.

The sets $\mathcal{N}$ and $\mathcal{D}$ are composed using the data introduced by~\citep{berglund2023reversal}. 
Both the names and descriptions in these sets were made up by GPT-4~\citep{openai2023gpt4}.
The fictitious data has not been encountered in the pretraining dataset of large language models. 
Consequently, we are able to simulate how these models acquire knowledge during pretraining and investigate the underlying causes of the reversal curse.
The training dataset consists of a total of 3,600 training samples. 
The test set comprises two tasks: one involves training with name-to-description data and testing the model using a paraphrased prompt, which we denote as the ``$N2D$'' task. 
The other task entails testing in the reverse order, where the description is given, and the model must generate the corresponding individual name. We refer to this as the ``$\mathop{N2D}\limits^{\longleftarrow}$'' task.
Each test task includes 300 test samples.
Figure~\ref{fig:data} demonstrates the training and test samples.

We select Llama-7B and 13B~\citep{touvron2023llama1}, the representative causal language models pre-trained with the NTP objective, along with GLM-2B and GLM-10B~\citep{DBLP:conf/acl/DuQLDQY022}, which support both ABI and NTP objectives, for investigating the impact of training objectives.
On the aforementioned fictitious dataset, we fine-tuned Llama models with the NTP objective, and GLMs with both NTP and ABI objectives, using the same settings as~\citep{berglund2023reversal}: a batch size of 4, a learning rate of 2e-5, and fine-tuning lasting for 10 epochs. 
Due to resource limitations, the models were fine-tuned using LoRA~\citep{hu2022lora} with $r=32$.
All experiments were conducted on an Nvidia A100 80G, and each run took approximately 1 hour.
Our default decoding strategy is greedy decoding.

We evaluate models' performance on two tasks using the Exact Match score~\citep{berglund2023reversal}, and the difference in accuracy between two tasks indicates the extent of the reversal curse.

\begin{table}[t]
    \centering
    \resizebox{0.9\linewidth}{!}{
    \begin{tabular}{l|c|cc}\toprule
         Model & Objective & N2D & $\mathop{N2D}\limits^{\longleftarrow}$ \\\midrule
        \multirow{2}*{GLM-2B} & NTP & 69.33 & 0.00 \\
        & ABI & 72.00 & 88.00 \\\midrule
        \multirow{2}*{GLM-10B} & NTP & 72.00 & 0.00 \\
        & ABI & 63.33 & 74.00 \\\midrule
        Llama-7B & NTP & 67.33 & 0.00 \\\midrule
        Llama-13B & NTP & 58.67 & 0.00 \\\bottomrule
    \end{tabular}
    }
    \caption{
    Models trained with NTP exhibit a more pronounced reversal curse when compared to the one trained for ABI (Llama does not support training with ABI).}
    \label{tab:glm-and-llama}
\end{table}

\subsection{NTP Exacerbates the Reversal Curse}
\label{sec:ntp-is-bad}
The experiment results, as shown in Table~\ref{tab:glm-and-llama}, reveal that GLM fine-tuned with the ABI objective exhibits resistance to the reversal curse. 
It maintains robust performance across two tasks. 
Their scores on the $N2D$ task, which involves generating long descriptions, are lower than those on the $\mathop{N2D}\limits^{\longleftarrow}$ task.
This is because the latter task requires generating short names and is relatively easier.
In contrast, both GLM and Llama models, trained with the NTP objective, demonstrate high precision on the $N2D$ task but experience a dramatic drop to zero when tackling the $\mathop{N2D}\limits^{\longleftarrow}$ task, revealing a severe reversal curse.

While these findings partially affirm our hypothesis, a crucial step remains to establish reliable evidence: the potential modification of Llama models to accommodate an ABI-like objective, enabling tokens to attend to both preceding and subsequent tokens during training. 
If, after fine-tuning, Llama models demonstrate relief from the reversal curse, we can confidently assert that training objectives indeed play a substantial role in the occurrence of the reversal curse.
Furthermore, as we confirm our hypothesis, the successful adaptation of Llama models to the ABI objective also contributes to mitigating the reversal curse. 
This is especially important in scenarios where models are fine-tuned with limited new data.

In the next section, we will present our approach to adapting Llama models for ABI-like objectives.

\begin{figure*}[t]
\vspace{3mm}
    \centering
    \includegraphics[width=\linewidth]{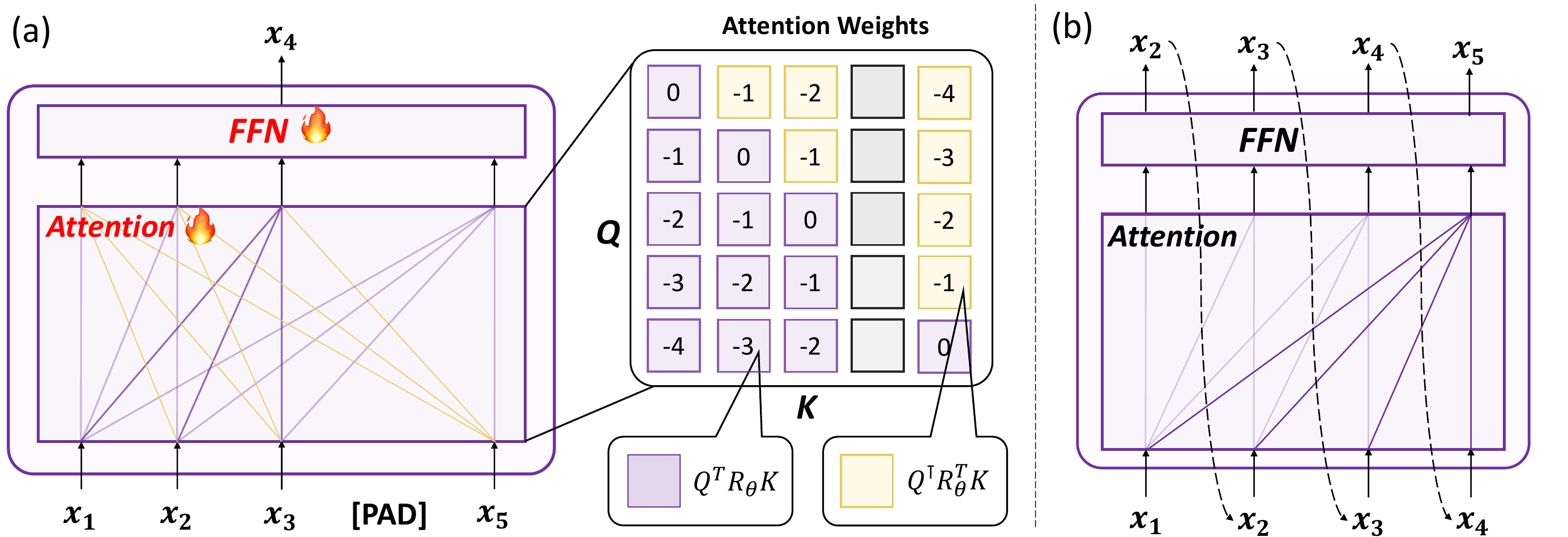}
    \caption{(a) Training details in \name. \name\ modifies the causal attention into a bidirectional one.
    Attention calculations are partitioned into two parts based on the relative positions of query and key vectors.
    Numbers in squares denote the relative distance between $q_{m}$ and $k_{n}$.
    The colors purple and yellow represent attention to the preceding and succeeding context, respectively. 
    Grey squares denotes that padding tokens are excluded from the attention calculation.
    (b) During inference, the language model adopts the causal attention as usual and predicts tokens autoregressively.
    For clarity, we only illustrate a single transformer layer and omit irrelevant modules.}
    \label{fig:overview}
\end{figure*}

\section{Adapting Llama Models for ABI-Like Objectives}

We present a novel fine-tuning framework that adapts the causal language models like Llama for an ABI-like objective. 
We name this framework as \textbf{BI}directional \textbf{C}ausal language model \textbf{O}ptimization (\name).
\name\ modifies the causal attention mechanisms during training ($\S$\ref{sec:bi-direction}) which ensures a seamless transition from unidirectional to fully bidirectional attention, thereby capturing the comprehensive contextual information from input data.
\name\ adopts an autoregressive blank infilling objective similar to GLM, with tailored modifications specifically designed for causal language models ($\S$\ref{sec:update}).
Figure~\ref{fig:overview} illustrates the overview of our approach and we delve into the details in below.

\subsection{Preliminary: Rotary Position Embedding}
When transitioning from a causal attention mechanism to a bidirectional one, addressing the out-of-distribution position information becomes crucial, and will be discussed in \S\ref{sec:bi-direction}.
We start by introducing the rotary position embedding implemented by Llama, as a necessary preliminary.

Rotary position embedding (RoPE,~\citealp{su2022roformer}) is a relative position embedding implemented during the attention calculation. 
When multiplying a query or key vector with a rotation matrix $R_{\theta}$, the position information is incorporated.
The $R_{\theta,m}$ is designed as a block diagonal matrix consisting of blocks with dimensions of $2\times2$, totaling $d/2$ such blocks. 
Specifically, the $i$-th block is defined as follows:
\begin{equation}
    \begin{aligned}
    R_{\theta_i, m} = \begin{bmatrix}
         \cos m \theta_{i} & -\sin m \theta_{i}\\
         \sin m \theta_{i} & \cos m \theta_{i}\\
    \end{bmatrix},
    \end{aligned}
\end{equation}
where $\theta_i := B^{-2i/d}, i\in [0,1,2,\dots,d/2-1]$ and $B$ is typically chosen as 10000.

With such a matrix design, the inner product of the query vector at $m$ position with the key vector at the $n$ position measures their relative distance:
\begin{equation}
\label{eq:qk-def}
    \begin{aligned}
        q_{m} & =R_{\theta,m} W_{q} x_{m},\quad k_{n}=R_{\theta,n} W_{k} x_{n},\\
        q^{\top}_{m} k_{n} &= (R_{\theta,m} W_{q} x_{m})^{\top} (R_{\theta,n} W_{k} x_{n}) \\
        & = (W_{q} x_{m})^{\top} R^{\top}_{\theta,m} R_{\theta,n} (W_{k} x_{n}) \\
        & = (W_{q} x_{m})^{\top} R_{\theta,n-m} (W_{k} x_{n}),
    \end{aligned}
\end{equation}
where $x_{m}$ and $x_{n}$ are $m$-th and $n$-th input of the current transformer layer; $W_q$ and $W_k$ project input hidden states to query and key vectors.

\subsection{Extending Causal Attention to Bi-Directional}
\label{sec:bi-direction}
Converting a unidirectional causal attention mechanism in a causal language model into a bidirectional one is non-trivial.
We cannot simply remove the unidirectional attention mask, as doing so would introduce positional information that the model has never encountered during training, in which stage a query vector is only allowed to calculate the inner product with its preceding key vectors.
This is evident in Eq.\ref{eq:qk-def}: the relative position $n-m$ is always non-positive during training but is positive when $q_m$ needs to attend to $k_{>m}$.
To address this issue, we propose a modification to the inner product between $q_m$ and $k_n$ for arbitrary values of $m$ and $n$ in a causal language model, as follows:
\begin{equation}
\label{eq:qk-extend}
\begin{aligned}
    q^{\top}_{m} k_{n}=\left\{
    \begin{aligned}
    &(W_{q} x_{m})^{\top} R_{\theta, n-m} (W_{k} x_{n}), n \leq m, \\
     &(W_{q} x_{m})^{\top} R_{\theta, m-n} (W_{k} x_{n}), n > m.\\
    \end{aligned}
    \right. 
\end{aligned}
\end{equation}
This adjustment ensures that when a query vector calculates an inner product with subsequent keys, there is no unexpected relative position information compared to training, as long as the relative distance between $m$ and $n$ does not exceed the maximum context length which is not within the scope of this paper. 

To implement the Eq.\ref{eq:qk-extend}: when $n \leq m$, we calculate the attention weights as usual;
In cases where $n > m$, we incorporate positional information with $R^{\top}_{\theta}$, the transposition of $R_{\theta}$.
Because $R^{\top}_{\theta,m}$ is equivalent to $R_{\theta,-m}$ for any given position $m$, we have:
\begin{equation}
\begin{aligned}
q^{\top}_{m} k_{n} &= (W_{q} x_{m})^{\top} (R^{\top}_{\theta,m})^{\top} R^{\top}_{\theta,n} (W_{k} x_{n}) \\
&= (W_{q} x_{m})^{\top} R_{\theta,m} R^{\top}_{\theta,n} (W_{k} x_{n}) \\
&= (W_{q} x_{m})^{\top} R^{\top}_{\theta,-m} R_{\theta,-n} (W_{k} x_{n}) \\
& = (W_{q} x_{m})^{\top} R_{\theta,m-n} (W_{k} x_{n}), \text{ when } n > m.
\end{aligned}
\end{equation}

Figure~\ref{fig:overview} illustrates this modification of attention calculation, where purple lines and squares denote that attention weights are calculated using the standard $R_{\theta}$ matrix, and yellow indicates that the query attends to its succeeding keys within the extended bidirectional attention mechanism. 
The annotated numbers indicate the relative distance between a query and a key vector, with all values being non-positive.

\subsection{ABI-like Objectives For Causal Language Models}
\label{sec:update}

\begin{figure*}[t]
\vspace{2mm}
    \centering
    \includegraphics[width=\linewidth]{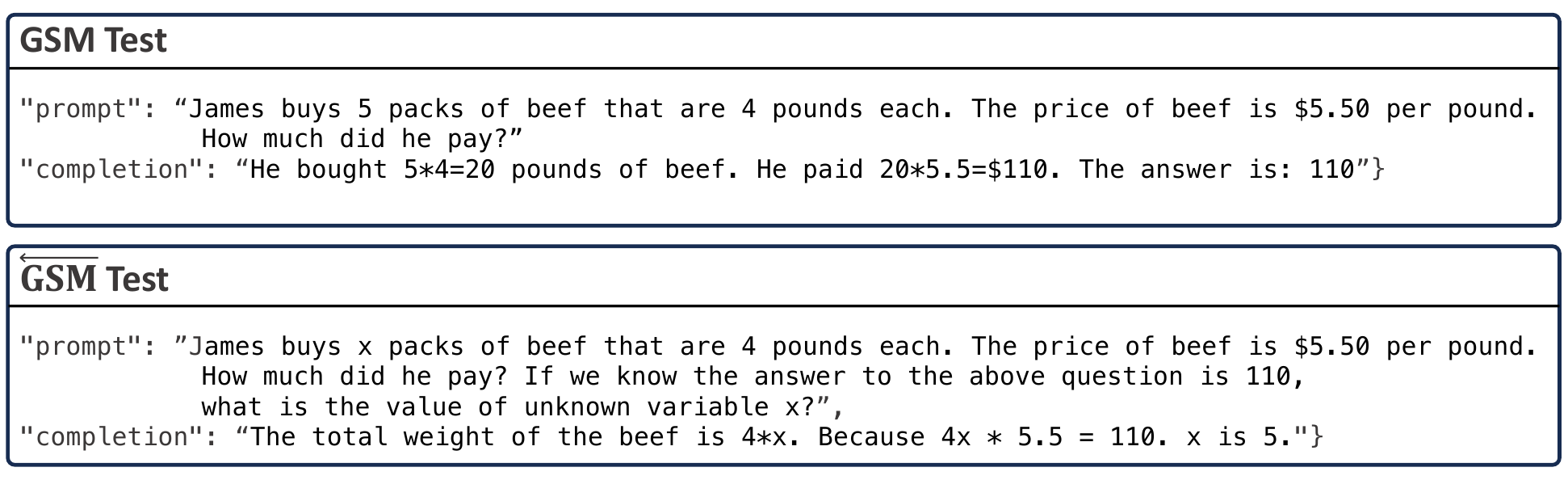}
    \caption{
    A test sample from the original GSM8k dataset~\cite{cobbe2021gsm8k}, alongside its ``reversal'' counterpart crafted by~\citet{yu2023metamath}.
    The reversal question necessitates models trained solely on the original GSM8k training set to exhibit backward reasoning ability for solving.
}
    \label{fig:math-data}
\end{figure*}

Based on the bidirectional attention, we make adjustments to the autoregressive mask denoising objective designed for causal language models like an autoencoding language model, thereby enabling a to-be-predicted token to have access to the entire context.
Our method incorporates several key components:

$\bullet$ During the training process, we randomly replace some tokens in input $X$ with a padding token, by a probability of $p_{M}$. 
In the text below, we use the default value of $p_{M}=0.15$ as it has been widely used as the mask token rate since BERT~\citep{devlin-etal-2019-bert}.
Considering that introducing a new mask token could create a gap between the training and inference, which impairs the performance~\citep{yang2020xlnet}, we choose the padding token instead of a \texttt{[MASK]} token, as causal language models typically lack a mask token in the vocabulary. 
The corrupted input $\hat{X}$ is then fed into the model.

$\bullet$ A padding token is excluded from attention calculations, meaning it is not attended to by other tokens, to prevent the introduction of semantic noise.
This is illustrated by grey squares in the attention weights in Figure~\ref{fig:overview}(a).

$\bullet$ At the $i-1$ output position, the model predicts the masked token at the $i$ input position, in alignment with the next-token prediction behavior during the model's pretraining stage.
Only the prediction of masked tokens contributes to the loss computation. 
Formally, the optimization objective is defined as follows:
\begin{equation}
    \mathop{\text{max}}\limits_{\Theta} \sum^{T-1}_{t=1} \mathbbm{1}(x_{t+1} \text{ is \texttt{[PAD]}}) \cdot \log p (x_{t+1} | \hat{X}; \Theta).
\label{eq:our-loss}
\end{equation}
Given that fine-tuning models solely with a mask denoising task could potentially diminish the model's proficiency in text continuation, we apply NTP objective in certain training steps to preserve the model's generative ability, denoting this fraction as \( p_O \). It is set as $0.5$ by default. 
We discuss the impact of \( p_M \) and  \( p_O \) in Section~\ref{sec:discuss}

The techniques described above introduce little gaps to causal language models.
Hence, a \name-tuned model can continue with the conventional inference process, i.e., autoregressively decoding the next token using a causal attention mechanism.

\section{Experiments and Analysis}
\label{sec:our-exp}

\subsection{Main Results}
We evaluate the efficacy of \name\ through two distinct tasks.
Initially, we apply \name\ to tackle the reversal curse encountered in fictitious name-to-descriptions task, as discussed in Section~\ref{sec:reason}. 
Subsequently, we evaluate its practical utility in solving mathematical problems, showcasing its capacity to improve the reversal reasoning skills of LLMs to a certain degree.

\begin{table}[t]
    \centering
    \resizebox{0.9\linewidth}{!}{
    \begin{tabular}{l|c|cc}\toprule
         Model & Objective & $N2D$ (EM) & $\mathop{N2D}\limits^{\longleftarrow}$ \\\midrule
        \multirow{2}*{Llama-7B} & NTP & 67.33 & 0.00 \\
        & \name & 69.67 & 68.33 \\\midrule
        \multirow{2}*{Llama-13B} & NTP & 58.67 & 0.00 \\
        & \name & 66.00 & 71.67 \\\bottomrule
    \end{tabular}
    }
    \caption{
    \name\ effectively mitigates the reversal curse during the fine-tuning of Llama with new knowledge, leading to significant enhancements in performance on the $\mathop{N2D}\limits^{\longleftarrow}$ task without any detrimental effects on the performance of the $N2D$ task. 
    Exact match scores are reported.}
    \label{tab:our-final-res}
\end{table}

\begin{table}[t]
    \centering
    \begin{tabular}{c|cc}\toprule
          Objective & $GSM$ & $\mathop{GSM}\limits^{\longleftarrow}$ \\\midrule
         NTP & 38.21 & 5.33 \\
        \name & 38.28 & 6.53 \\\bottomrule
    \end{tabular}
    \caption{We fine-tune a Llama-7B model using the GSM8k dataset~\cite{cobbe2021gsm8k} with NTP and \textit{\name}, respectively. 
    The averaged answer accuracy is reported.
    The tuned models are evaluated on the original test questions (denoted as $GSM$) and the reversal questions constructed by \citet{yu2023metamath} (denoted as $\mathop{GSM}\limits^{\longleftarrow}$).}
    \label{tab:our-math-res}
\end{table}

\paragraph{Mapping Fictitious Names to Descriptions}
We use \name\ for fine-tuning Llama models (7B and 13B) and evaluating performance on both the $N2D$ and $\mathop{N2D}\limits^{\longleftarrow}$ tasks (See Figure~\ref{fig:data} for data details). 
Experimental results are presented in Table~\ref{tab:our-final-res}. 
Upon applying our proposed \name, a significant accuracy enhancement is observed for the reverse task, rising from 0\% to around 70\%. 
Building on this observed increase, we complete the final step outlined in \S\ref{sec:ntp-is-bad} to validate our hypothesis that training objectives can indeed influence the reversal curse, serving as one of its contributing factors. 
Moreover, \name\ itself contributes as a fine-tuning method that does not introduce extra reversal curse for causal language models.

\paragraph{Backward Math Solving}
To further showcase \name's effectiveness, we tackle a more practical task: math solving.
The basic experimental setup involves teaching LLMs basic math-solving skills and testing them with ``reversed logic'' approaches to address math problems.
The dataset~\cite{yu2023metamath}, derived from GSM8k~\cite{cobbe2021gsm8k}, features math questions reversed compared to those in the original GSM8k dataset. 
Figure~\ref{fig:math-data} illustrates a data point from GSM8k alongside its reversed counterpart. 
We use NTP and \name\ to fine-tune a Llama-7B model on the original GSM8k dataset, respectively. We assessed the performance of tuned models on both the original and reversed test sets.
All training hyperparameters and configurations follow~\citet{yu2023metamath}. 
The results are shown in Table~\ref{tab:our-math-res}.

We observe that \name\ maintains its performance on the original test set while achieving an increase of more than 1 point in solving reversal math problems, which is a decent improvement in math solving task and passes the t-test with p-value < 0.01.
Given that models haven't seen any backward reasoning chains for solving these types of math questions during fine-tuning, the improvement can be attributed to \name, showcasing its capability to enhance causal language models with a more versatile reasoning ability.

\paragraph{Translation}
The dual nature of translation data is particularly well-suited for evaluating the reversal curse. 
We assume that given the "language $X$-language $Y$" order training data, the model may perform poorly in "language $Y$-to-language $X$" translation due to the reversal curse. 
To demonstrate this, we conducted a simple machine translation task from Chinese to English. 
Here are the experimental details:

\begin{CJK*}{UTF8}{gbsn}
We developed a set of Chinese-to-English translation examples structured as follows: ``When translating the Chinese term `另外一个' into English, the equivalent expression is `Another one'.''
For testing, we reversed the training data to conduct English-to-Chinese translation tasks, such as: ``When translating the English phrase `another one' into Chinese, the corresponding Chinese expression is,'' with the correct response being `另外一个.'
\end{CJK*}
We tested this task on the Llama-7b model, which has limited Chinese language capacity. 
Both the training and testing datasets comprise 100 examples each. 
We assessed the translation performance using the Exact-match metric.
The experiment results are listed in Table~\ref{tab:trans}.
Note that the 0-shot accuracy for Llama is 50\%, reflecting its original bilingual abilities. 
BICO outperforms NTP by 6\%, highlighting its potential effectiveness in real-world tasks such as machine translation.

\begin{table}[t]
\vspace{2mm}
    \centering
    \begin{tabular}{l|c|c|c}
    \toprule
         & 0-shot & NTP & \name \\\hline
        EM (\%) & 51 & 63 & 69 \\\bottomrule
    \end{tabular}
    \caption{\name\ enhances the utility of training data of the reverse-translation task, thereby improving the accuracy of reverse translation. 
    Note that the Llama model has inherent multilingual ability, shown in the zero-shot score.
    }
    \label{tab:trans}
\end{table}

\section{Discussions}
\label{sec:discuss}
This section includes some necessary analysis about our proposed method \name, as well as contemporary research on the reversal curse.

\subsection{Method Analysis}
The configuration of hyperparameters in \name, including $p_M$ and $p_O$, adheres to conventional practices akin to BERT~\cite{devlin-etal-2019-bert}.
However, in BERT, the mask token ratio $p_M$ and the trade-off between two pretraining tasks $p_O$ lack discussion and explicit formalization. 
We have studied the impact of our hyperparameters and details can be found in the appendix~\ref{apx:analysis}.

\subsection{The Mitigation of Reversal Curse}

Some contemporary works aim to tackle the reversal curse through knowledge editing or data augmentation.

Knowledge editing~\cite{ma2023untying} integrates reverse knowledge directly into model parameters, ensuring performance. 
However, this method is labor-intensive, requiring meticulous annotation and complex implementation. 
It necessitates sentences to adhere to a specific structure of one subject, followed by a relation, and then one object. 
In contrast, \name\ is agnostic to sentence structures and easier to implement.

Following us,~\citet{guo2024mitigating} propose to mitigate the reversal curse through data augmentation.
It can generate explicit reversal data for training, potentially enhancing performance in complex scenarios. 
However, this method may suffer from label leakage issues in research experiments, and thus we do not compare \name\ with it.

Here is a scenario that both \name\ and ABI fail to mitigate the reversal curse but data augmentation can solve according to~\cite{guo2024mitigating}.
Consider the inverse relation of the previously studied $R_{\mathcal{N}2\mathcal{D}}$, denoted as $R_{\mathcal{D}2\mathcal{N}}$. 
The mapping from $\mathcal{D}$ to $\mathcal{N}$ also forms a bijection:
$R_{\mathcal{D}2\mathcal{N}} = \{<d, n> | \ d \ \text{describes}\ n, \ (n \in \mathcal{N}) \ \land \ (d \in \mathcal{D})\}$.

\begin{figure}
    \centering
    \includegraphics[width=0.9\linewidth]{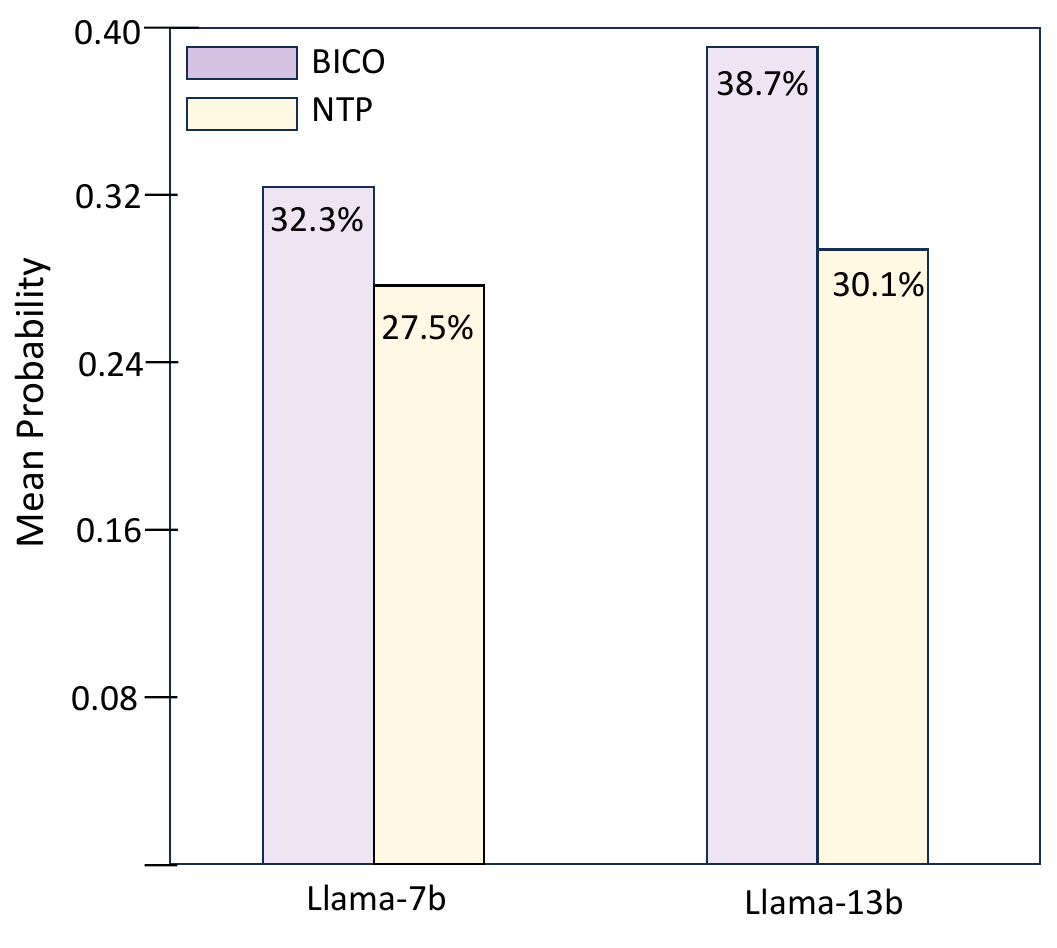}
    \caption{The probability of the desired completion given prompts provided by various models in $\mathop{D2N}\limits^{\longleftarrow}$ task. 
This probability is evaluated across the entire test set and is presented as an average. 
It is clear that \name\ enhances the likelihood of achieving ground truth prediction.}
    \label{fig:logp-results}
\end{figure}

All experimental setups and training details remain identical to those previously introduced.
Figure~\ref{fig:d2n-data} in the appendix illustrates the data construction associated with this relationship.
After tuning, we encounter an preplexing phenomenon: 
As shown in Figure~\ref{fig:logp-results}, after tuning, we observe a perplexing phenomenon: while \name\ improves the likelihood of ground truth compared to NTP-tuned models, its impact on exact match or BLEU scores~\cite{papineni-etal-2002-bleu} is insignificant (refer to appendix~\ref{apx:analysis} for likelihood computation and detailed results).
We have scrutinized the decoding strategies, including beam search, top-k, and top-p sampling, but they do not exhibit much difference.

One plausible hypothesis is that pre-trained LLMs may have developed a ``common sense'' during pre-training.
This ``common sense'' suggests that, compared with description-to-name, a name-to-description relationship is more inclined to be a one-to-many mapping, resulting in confusion in the $\mathop{D2N}\limits^{\longleftarrow}$ task.
For instance, when posed with the question, ``Who is Joe Biden?'', the model might respond with many correct descriptions such as ``a car enthusiast,'' or ``an elderly man,'' rather than the ``American president.''
While explicit data augmentation may offer a solution to this issue, supported by findings in recent interpretability research~\cite{wang2024grokked}, this method merely encourages the model to memorize reversed data rather than enhancing its reasoning abilities.

\section{Conclusions}
We are the first to study the underlying causes of the reversal curse and attribute it to a combination of training objectives and certain inference mechanisms. 
When examining the impact of training objectives, we introduce an innovative fine-tuning approach for causal language models named \name. 
This method customizes Llama models for ABI-like objectives, thereby mitigating the reversal curse that emerges during the training phase. 
We hope to draw the community's attention to the prevalent configuration of large language models, especially highlighting the inherent limitations in the existing training paradigm.

\section*{Limitations}
As limitations, there remain some open research questions that warrant further investigation:
Firstly, quantifying the influence of other processes on advanced models, such as RLHF, on the reversal curse poses a more intricate challenge and needs more study.
Secondly, understanding the various mechanisms beyond the training objective that contribute to exacerbating the reversal curse is crucial, and we defer this task to future research endeavors.

\section*{Acknowledgements}
This work was supported by the National Natural Science Foundation of China (NSFC Grant No. 62122089), Beijing Outstanding Young Scientist Program NO. BJJWZYJH012019100020098, and Intelligent Social Governance Platform, Major Innovation \& Planning Interdisciplinary Platform for the ``Double-First Class'' Initiative, Renmin University of China, the Fundamental Research Funds for the Central Universities, and the Research Funds of Renmin University of China.
Ang Lv is supported by the Outstanding Innovative Talents Cultivation Funded Programs 2024 of Renmin University of China.

\newpage
\bibliography{example_paper}
\bibliographystyle{acl_natbib}

\newpage
\appendix
\section{Method Analysis of \name}
\label{apx:analysis}

We investigate the impact of hyperparameters in \name\ on mitigating the reversal curse induced by training objectives. 
It's important to note that we do not aim at searching for the best performance for a particular task, but rather at studying the characteristics of \name.
The model we used in this section is Llama-7B.

\paragraph{The Choice of $p_O$}
The original ABI uses special tokens like "$<\texttt{BOS}>$" and "$<\texttt{EOS}>$" to indicate the start and the end of masked sequence. 
However, it is worth noting that this differs from the manner in which Llama makes use of these special tokens.
Therefore, we have opted not to utilize them as markers for masking in \name.
As a result, we find Llama models tuned with \name\ encounter challenges in effectively terminating generation.
We observed that they tend to produce lengthy, topic-related descriptions. 
To solve this issue, we introduce the solution: at each training step, the model is optimized with the NTP objective with probability $p_O$ and is optimized with the autoregressive mask denoising with probability $1 - p_O$.

\begin{figure}[t]
    \centering
    \includegraphics[width=\linewidth]{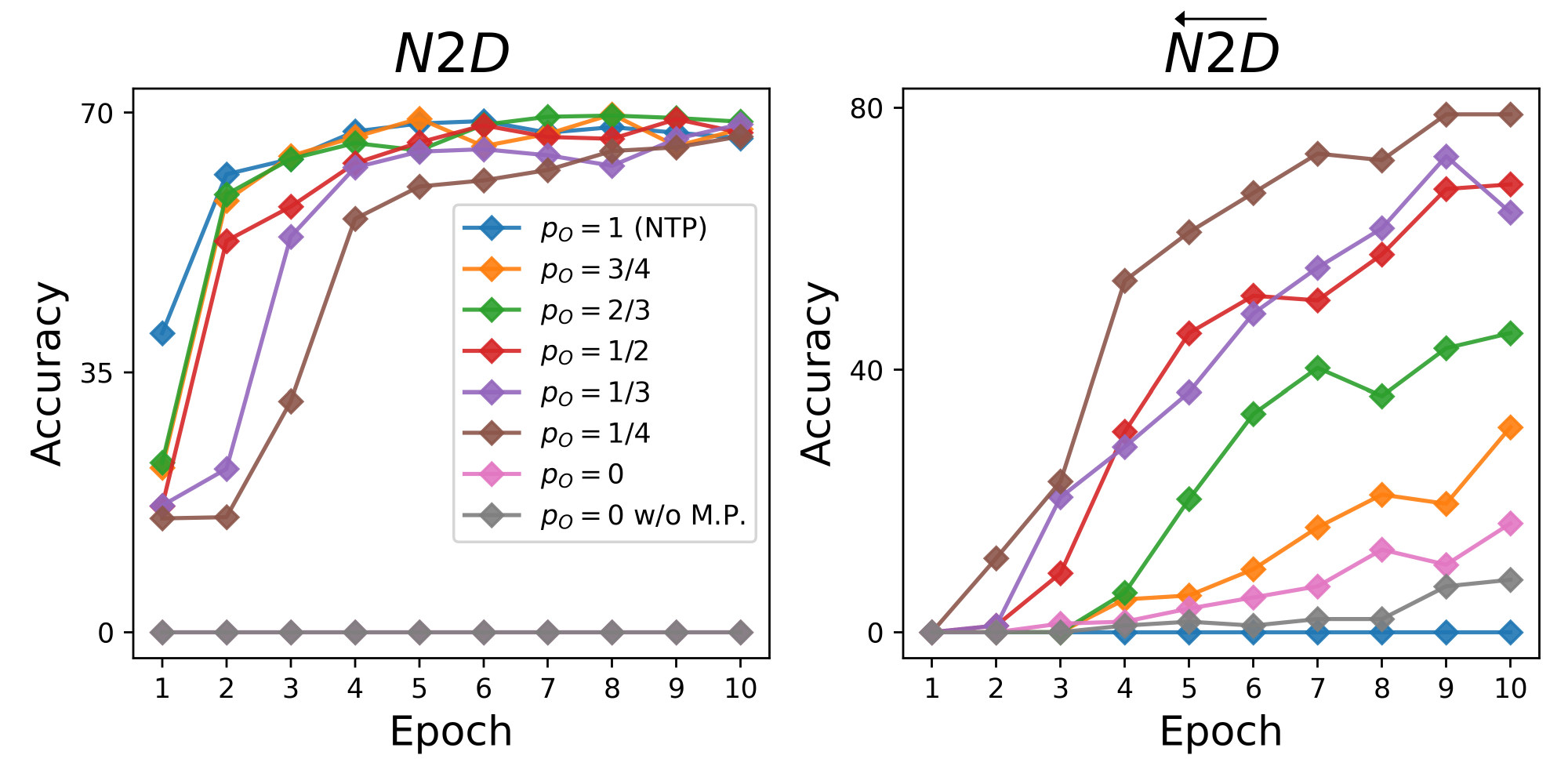}
    \caption{\(p_O\) balances the model's comprehensive understanding of training data and generative ability. 
    When the appropriate \(p_O\) value is selected, \name\ mitigates the reversal curse in the $\mathop{N2D}\limits^{\longleftarrow}$ task.
    Absence of modified position embeddings (w/o M.P.) impedes the learning process and results in inferior outcomes due to the out-of-distribution problem.}
    \label{fig:our-res}
\end{figure}

\begin{figure}[t]
    \centering
    \includegraphics[width=\linewidth]{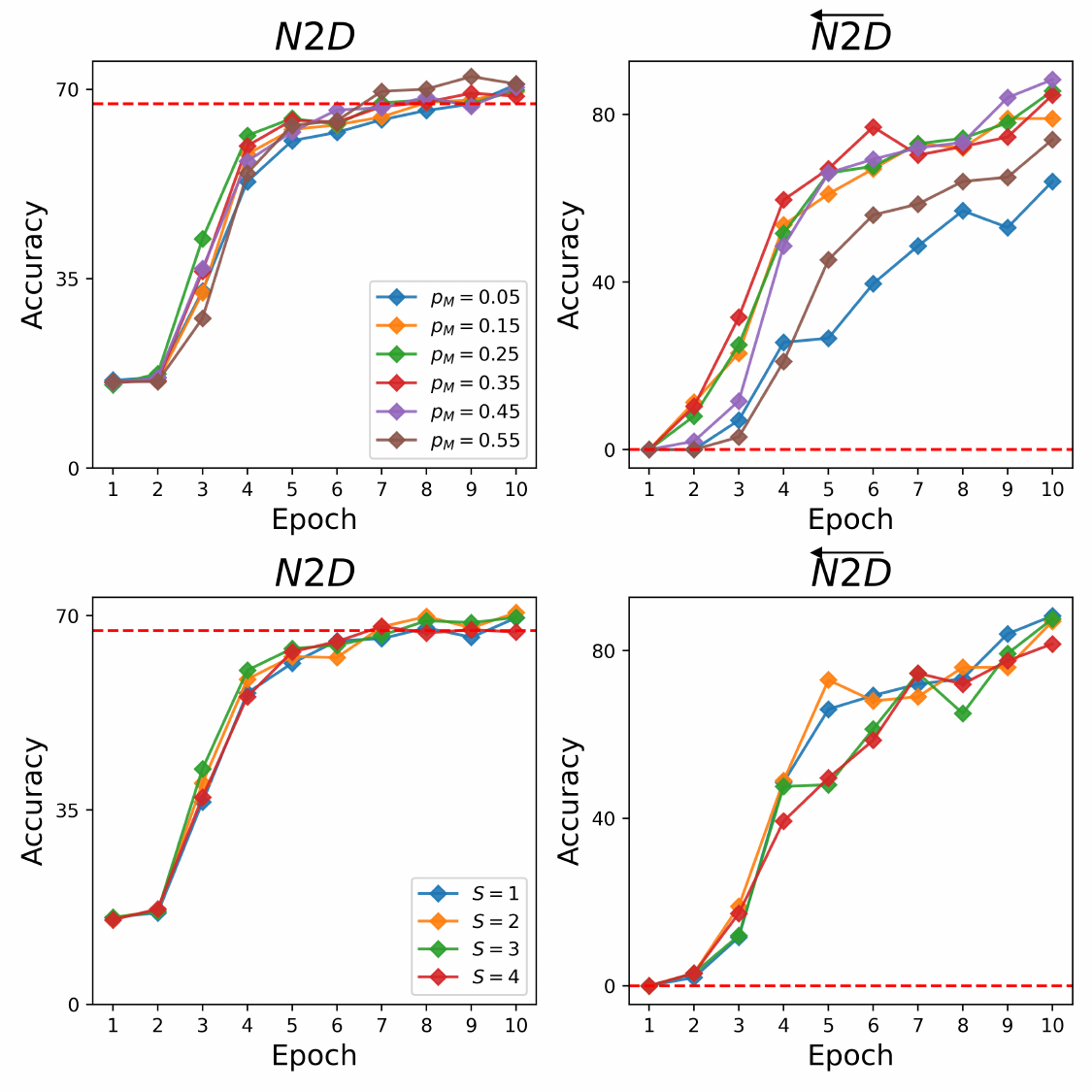}
    \caption{The impact of mask strategy on model performance.
    From the upper portion in the figure, the model performance remains consistent across a wide range of $p_M$ values (0.15 to 0.45).
    We also find that the span masking and i.i.d. masking do not exhibit notable differences.
    For the ease of comparison, we use red dashed line to denote the best results provided by fully NTP-tuned Llama.}
    \label{fig:our-mask}
\end{figure}

\begin{figure*}[t]
    \centering
    \includegraphics[width=\linewidth]{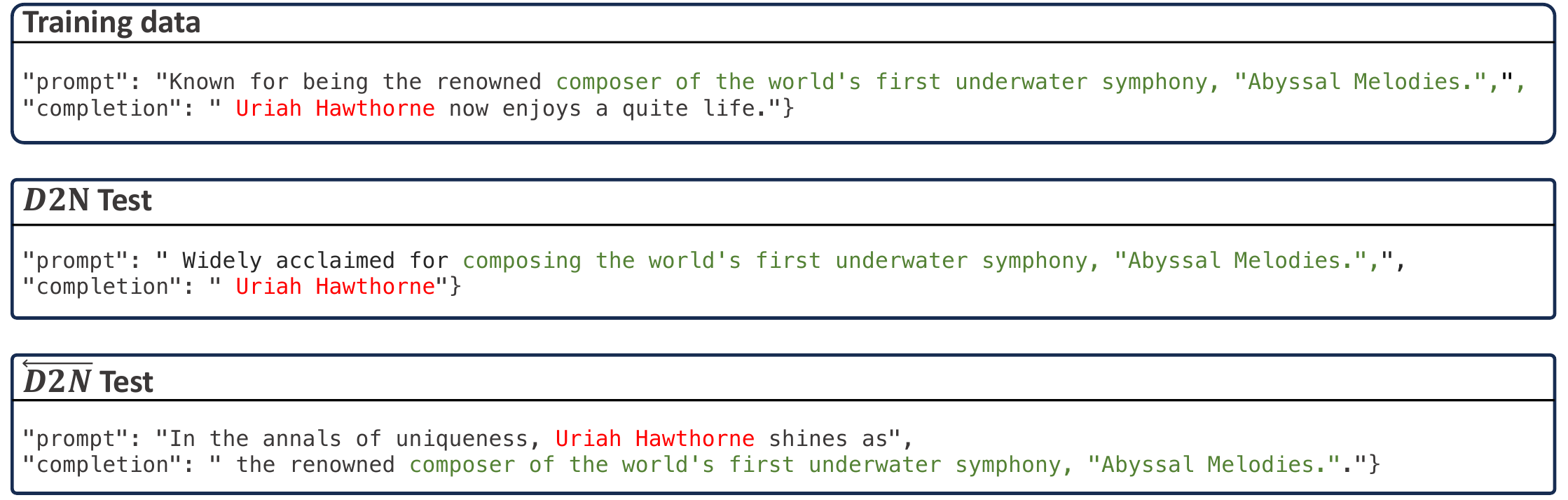}
    \caption{
Data employed for studying the reversal curse on relation $R_{D2N}$. 
All names and descriptions are fictitious.
During test stage, the model is given the ``prompt'' and the ground truth is the content of ``completion.''}
    \label{fig:d2n-data}
\end{figure*}

We study the choice of $p_O$, while maintaining a constant mask rate of $p_M=0.15$, a widely adopted value in autoencoding models~\citep{devlin-etal-2019-bert,2020t5}.
Our results are illustrated in Figure~\ref{fig:our-res}.
We observed that because of the previously discussed issue, when $p_O=0$, models can not generate accurate description in $N2D$ task, and performs poor in $\mathop{N2D}\limits^{\longleftarrow}$ task.
The model fully fine-tuned with NTP ($p_O = 1$) suffers from the reversal curse, achieving an accuracy rate of 0$\%$ in $\mathop{N2D}\limits^{\longleftarrow}$ task.

A balanced outcome is achieved with a small \( p_O \) around 1/4, enabling the model to preserve its generative ability while thoroughly learning from the training data. 
Consequently, there is a remarkable improvement in N2D-reverse task performance, soaring from 0$\%$ to around 80$\%$, with sustained proficiency in the $N2D$ task.
We also explore the impact of fine-tuning the model without modifying its attention calculation ($p_O=0$, w/o M.P.) to address out-of-distribution positions.
Due to the need for a portion of the parameter updates to tackle OOD issues in position embeddings~\citep{chen2023extending}, the learning process is slowed down.

\paragraph{The Mask Strategy}
We investigate the mask strategy in \name. 
We set the parameter $p_O$ at a constant value of 1/4 and vary the mask probability $p_M$ from 0.05 to 0.55, increasing in increments of 0.1. 
The results are illustrated in the upper portion of Figure~\ref{fig:our-mask}. 
It was observed that extreme values of $p_M$, such as 0.55 or 0.05, yielded suboptimal results, while intermediate values did not show significant divergence. 

Additionally, we explore the mask span. 
Previous studies~\cite{2020t5,joshi-etal-2020-spanbert} suggest that masking a contiguous span of tokens can be more effective than employing independently and identically distributed (i.i.d.) mask tokens, equivalent to a mask span of 1. 
In our experiment, we explored mask span length $S$, and the results are in the bottom of Figure~\ref{fig:our-mask}. 
We did not observe any clear performance differences among different $S$ settings.

\section{$D2N$ and $\mathop{D2N}\limits^{\longleftarrow}$ Tasks}

Figure~\ref{fig:d2n-data} illustrates a data point in $D2N$ and $\mathop{D2N}\limits^{\longleftarrow}$ tasks.

For specifics regarding likelihood computation in Figure~\ref{fig:logp-results}: 
We computed the likelihood of the ground truth assigned by LLMs after fine-tuning as follows:
\begin{equation}
\begin{aligned}
&p(\text{completion}|\text{prompt}) = e^{-\mathcal{L}_{NLL}},\\
& \text{where } \mathcal{L}_{NLL} = -\sum_{i=k}^{l} \log p(t_{i} | t_{0:i-1}).
\end{aligned}
\end{equation}
Here, $t_{i}$ denotes the $i$-th token within the sequence, which has a length of $l$.
Importantly, we do not take into account the positions corresponding to the prompt (the first $k$ tokens) when computing the loss.

The Exact Match scores and BLEU~\cite{papineni-etal-2002-bleu} scores (specifically for the $\mathop{D2N}\limits^{\longleftarrow}$ task, which involves generating long descriptions) of models tuned by different training objectives are reported in Table~\ref{tab:glm-and-llama-d2n}.

\begin{table}[t]
    \centering
    \caption{
    Models tuned by different training objectives consistently struggle in $D2N$ and $\mathop{D2N}\limits^{\longleftarrow}$ task.}
    \resizebox{\linewidth}{!}{
    \begin{tabular}{l|c|ccc}\toprule
         Model & Objective & D2N (EM) & $\mathop{D2N}\limits^{\longleftarrow}$ (EM) & $\mathop{D2N}\limits^{\longleftarrow}$ (BLEU) \\\midrule
        \multirow{2}*{GLM-2B} & NTP & 100.00 & 0.00 & 19.70 \\
        & ABI & 100.00 & 0.07 & 22.13 \\\midrule
        \multirow{2}*{GLM-10B} & NTP & 100.00 & 0.00 & 19.01 \\
        & ABI & 99.33 & 1.67 & 22.15 \\\midrule
        \multirow{2}*{Llama-7B} & NTP & 100.00 & 0.00 & 19.65 \\
        & \name & 99.67 & 1.00 & 21.00 \\\midrule
        \multirow{2}*{Llama-13B} & NTP & 98.67 & 0.00 & 20.62 \\
        & \name & 99.33 & 1.33 & 22.15 \\\bottomrule
    \end{tabular}}
    \label{tab:glm-and-llama-d2n}
\end{table}


\end{document}